\documentclass{article}
\usepackage{spconf,amsmath,epsfig}
\usepackage{color}
\usepackage{pifont}
\begin{document}\sloppy

\def\x{{\mathbf x}}
\def\L{{\cal L}}




\title{Context-constrained Accurate Contour Extraction for Occlusion Edge Detection}

%
\name{Rui Lu$^{1}$, Menghan Zhou$^{1}$, Anlong Ming$^{1}$ and Yu Zhou$^{2,*}$}

\address{$^{1}$Beijing University of Posts and Telecommunications, Beijing, China\\
$^{2}$Huazhong University of Science and Technology, Wuhan, China\\
{\tt\small {\{lurui,zhoumenghan,anlongming\}@bupt.edu.cn} yuzhou@hust.edu.cn}
}

\maketitle

\renewcommand{\thefootnote}{}
\footnotetext{
This work was supported by the National Natural Science Foundation of China 61703049.\\
$*$ Corresponding Author.
}

\begin{abstract}

Occlusion edge detection requires both accurate locations and context constraints of the contour. Existing CNN-based pipeline does not utilize adaptive methods to filter the noise introduced by low-level features. To address this dilemma, we propose a novel Context-constrained accurate Contour Extraction Network (CCENet). Spatial details are retained and contour-sensitive context is augmented through two extraction blocks, respectively. Then, an elaborately designed fusion module is available to integrate features, which plays a complementary role to restore details and remove clutter. Weight response of attention mechanism is eventually utilized to enhance occluded contours and suppress noise. The proposed CCENet significantly surpasses state-of-the-art methods on PIOD and BSDS ownership dataset of object edge detection and occlusion orientation detection.

\end{abstract}
\section{Introduction}
\label{sec:intro}
\begin{figure}[t]
\centerline{\psfig{figure=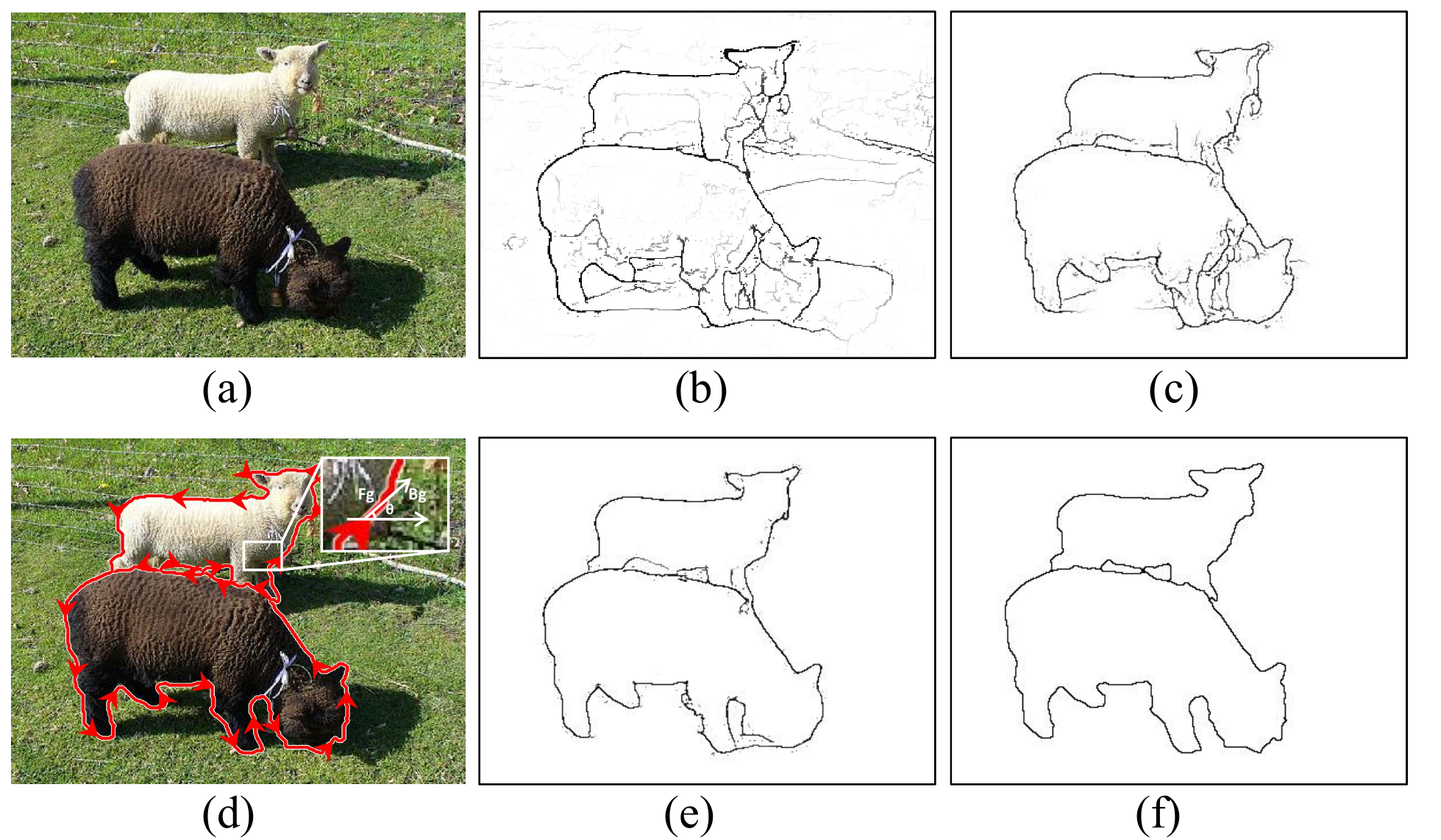,width=3.4in}}
\caption{
(a) an input image, (b) object edge map of DOC-HED, (c) object edge map of DOOBNet, (d) occlusion boundaries (the red arrows) represented by orientation $\theta\in(-\pi,\pi]$ (tangent direction of the edge), using the "left" rule where the left side of the arrows means foreground, (e) object edge map of ours, (f) object edge map of ground truth.
}
\label{Fig:demo}
\end{figure}

Occlusion edge detection is  
significant
in visual tracking, mobile robot, and other related fields \cite{Zhou2012NIPS,Teo2015Fast,Yu2016Human,Yu2016Similarity}.
In a monocular image, occlusion edges satisfy the depth discontinuity between regions, which represent the relationship between foreground and background from the view of observer. As shown in Fig.\ref{Fig:demo}(d), two sheep occlude the lawn, so their contours are regarded as occlusion edges, while their shadows are not.

During the past few years, several methods have been proposed \cite{Ren2006Figure,Zhou2014ONLINE,Zhou2018Learning}, which mainly apply hand-crafted features to machine learning skills.
Zhou et al. \cite{inproceedings} design a double layer model to address prediction drifts to the non-object backgrounds.
Ma et al. \cite{Ma2017Object} extract feature samples and apply the kernel ridge regression to acquire occlusion edge map. 
Recently, CNN 
is widely used 
to exploit local and non-local cues \cite{Long_2015_CVPR}, and has been successfully applied in visual tasks such as edge detection \cite{Xie2016Holistically} and semantic segmentation \cite{Chen2016DeepLab}. 


Previous methods {\cite{Wang2016DOC,Wang2018DOOBNet}} use a continuous occlusion orientation map and a binary object edge map
(Fig.\ref{Fig:demo}(d)(f))
to predict occlusion relationship and boundaries between objects, respectively.
Spatial information and large receptive field directly affect the accuracy.
However, few works 
manage to
extract affluent multi-level features and fuse them appropriately, resulting in noise for the prediction.
DOC \cite{Wang2016DOC} simply concatenates multi-level features,
causing the noise from low layers to be preserved. As shown in Fig.\ref{Fig:demo}(b),
numerous false positive non-occluded pixels and external texture are detected. DOOBNet \cite{Wang2018DOOBNet}
obtains low-level features directly from the original image, which contains massive noise. As seen in Fig.\ref{Fig:demo}(c),
there exists plenty of missed inspection.

\begin{figure*}[t]
\centerline{\psfig{figure=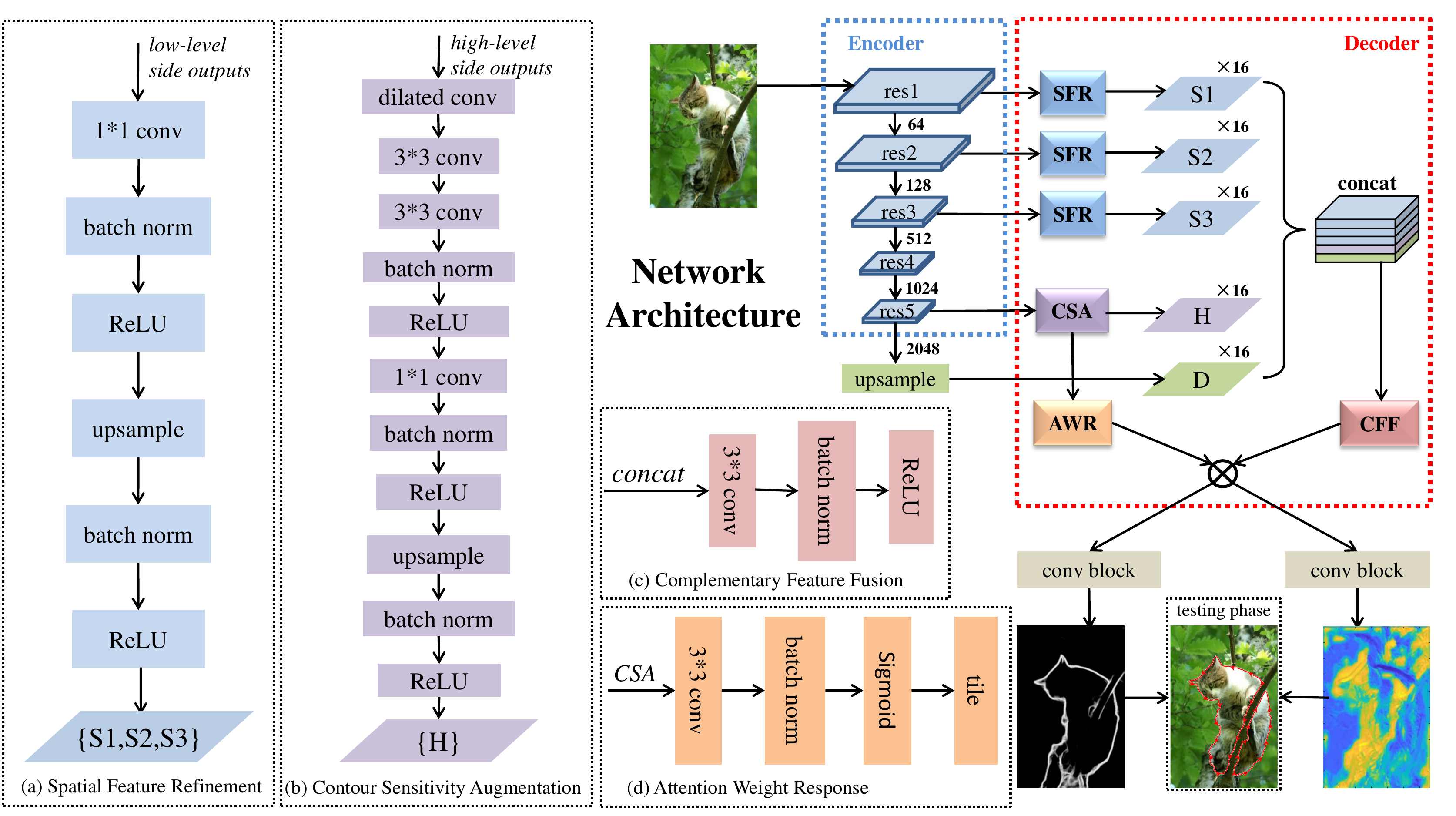,width=6.5in}}
\caption{
Illustration of our CCENet architecture. (a) Spatial Feature Refinement. (b) Contour Sensitivity Augmentation. (c) Complementary Feature Fusion. (d) Attention Weight Response. $\otimes$ represents the element-wise product operation.
}
\label{Fig:network}
\end{figure*}

In this paper, we attempt to improve the accuracy of edge localization, reinforce the intensity of occluded contours and suppress the noise of trivial edges. Aiming to acquire details of the edge positioning, we exploit the side outputs of low-level layers. In order to enhance the contours of an object, we make full use of the high-level side output layers which retain large receptive field. Our method manages to complement feature maps from various levels, thereby improving the model to
restore detailed location of contours and remove non-occluded pixels. To further enhance the ability to acquire the contours, we utilize the weight response of attention mechanism to distinguish the importance of pixels at different positions,
which guides the network to forward in a direction with clear contour and less noise around it.
As shown in Fig.\ref{Fig:demo}(e), our predicted edge map contains clear contours and little surrounded noise. Besides, it perceives precise edge positioning, which is almost consistent with the ground truth.
We validate the proposed method through abundant experiments on PIOD \cite{Wang2016DOC} and BSDS ownership dataset \cite{Ren2006Figure}. Superior results demonstrate the effectiveness of our approach.

The main contributions of this paper are as follows:
\begin{itemize}
  
    \item We propose a novel method to utilize Spatial Feature Refinement (\textbf{SFR}) and Contour Sensitivity Augmentation (\textbf{CSA}) to acquire low-level and high-level information.
    \item We design two specific modules: Complementary Feature Fusion (\textbf{CFF}) and Attention Weight Response (\textbf{AWR}) to fuse feature maps from different levels and take advantage of contour context information supplied by \textbf{CSA}.
    
\end{itemize}

\section{Background}

In this section, the representation of object edge map and occlusion orientation map are introduced. 


\subsection{Occlusion Edge Representation}

Mathematically, the $n^{th}$ object edge map is denoted by ${{E}_{n}}=\{{{e}_{i}}^{(n)}\}_{i=1}^{\left| {{E}_{n}} \right|}$ and the $n^{th}$ occlusion orientation map is denoted by ${{A}_{n}}=\{{{a}_{i}}^{(n)}\}_{i=1}^{\left| {{A}_{n}} \right|}$. Both maps have the same size as the input original image. Notably, $e_i^{(n)}\in \{0,1\}$ and $a_i^{(n)}\in (-\pi ,\pi ] $ denote the ground truth binary value and continuous value of the image, respectively. 
We use ${{\overline{E}}_{n}}$ and ${{\overline{A}}_{n}}$ to represent the corresponding predicted value of ${{E}_{n}}$ and ${{A}_{n}}$. Specifically, if $\overline{e}_{i}=1$ at pixel $i$, ${{\overline{a}}_{i}}$ means the tangent of the edge. Under the "left" rule, its direction is used to represent the occlusion relationship. While if $\overline{e}_{i}=0$ at pixel $i$, $\overline{a}_{i}$ is meaningless.


\subsection{Loss Function}


Following \cite{Wang2018DOOBNet}, the loss function for a batch of images is formulated as:
\begin{equation}
\label{Eq:loss_all}
l(W)=\frac{1}{M}(\sum\limits_{j}{\sum\limits_{i}{AL({\overline{e}_{i}},{{{e}}_{i}})+\lambda \sum\limits_{j}{\sum\limits_{i}{SL(f({\overline{a}_{i}},{{{a}}_{i}})}}}})
\end{equation}
where $W$ is the collection of all standard network layer parameters
, $M$ is mini-batch size, $j$ is the $j^{th}$ image in $M$. $AL$ and $SL$ 
are
the Attention Loss and the Smooth $L_1$ Loss, respectively.





\section{The CCENet}

In this section, the novel Context-constrained accurate Contour Extraction Network (CCENet) is proposed. Fig.\ref{Fig:network} shows the pipeline of the proposed CCENet. For the original RGB image, Spatial Feature Refinement (\textbf{SFR}) and Contour Sensitivity Augmentation (\textbf{CSA}) are exploited to obtain low-level and high-level feature maps, respectively (see Sec.\ref{sec:SFR} and Sec.\ref{sec:CSA}). In order to fuse the features from different levels, a Complementary Feature Fusion (\textbf{CFF}) is designed (see Sec.\ref{sec:CFF}). Besides, Attention Weight Response (\textbf{AWR}) is learned from end-to-end, in order to restrict the model to the part reliable for occlusion edge detection (see Sec.\ref{sec:AWR}).





\subsection{Spatial Feature Refinement}
\label{sec:SFR}

For occlusion edge detection, spatial features extracted by the early layers retain more fine-grained spatial details and thus are useful for precise localization. However, due to the pooling operators within the network backbone, the latter layers are less effective to locate the occlusion edge because of the gradual reduction of spatial resolution. 


We alleviate this issue by using the first three layers of side outputs of the $Res50$ \cite{He_2016_CVPR} in the encoder phase. They encode rich spatial details without losing the details of edge locations,
owing that the feature maps have the similar resolution with the original image. However, different from DOOBNet, which simply utilizes plain convolution blocks to obtain low semantic information directly from the original image, we employ three corresponding side outputs from the first three layers. It is effective to avoid introducing noise of massive color, texture, and gradient features, which are usually contained in the original image. 
A convolution block is exploited to reduce the number of channels and extract spatial features. Besides, a batch normalization layer \cite{Ioffe2015Batch} and a $ReLU$ are utilized to normalize the range. 
Each side output is resized by the bilinear upsampling. The spatial-refined maps, $i.e.,\{S_1,S_2,S_3\}$, which have the same size as the original image, are thus obtained. 



\subsection{Contour Sensitivity Augmentation}
\label{sec:CSA}

Along with the CNN forward propagation, 
the semantic discrimination between contours from different
categories is strengthened. Meanwhile, response to trivial pixels is gradually weakened.
On this account, to capture richer context information for the perception of contours, extracting features from a larger region is considered.

\begin{figure}[t]
\centerline{\psfig{figure=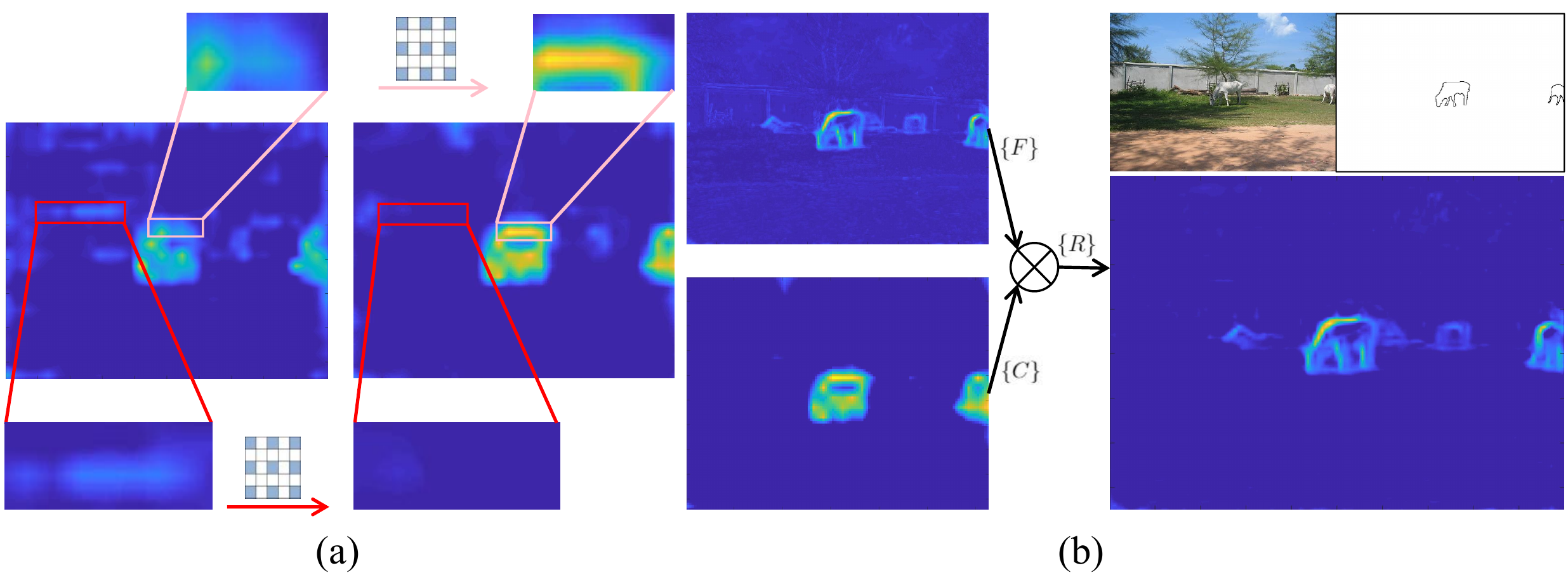,width=3.2in}}
\caption{
Feature maps demos. (a) effect of the dilated convolution in \textbf{CSA}, (b) effect of the weighted operation in \textbf{AWR}.
}
\label{Fig:modules}
\end{figure}

For the purpose of enlarging the receptive field of feature maps while not decreasing resolution, 
a pooling layer after $res5$ is replaced by
a dilated convolution \cite{Yu2017Dilated} with parameter of 2. The dilated convolution is utilized to form the constraints between pixels on the contour. 
In Fig.\ref{Fig:modules}(a), pixels within the pink box, which ought to be considered as the contour, are surrounded by numerous pixels 
which belong 
to continuous boundaries. As context features, these pieces of information are sensitive to the whole contour-level detection, resulting in a unified global edge enhancement of the pixels within the pink box.
Meanwhile, surrounding the miscellaneous pixels within the red box, there are few boundaries, making the pixels incapable to be recognized as the part of a contour. In other words, there is almost no integrated edge around the clutter, so it will not be recognized as the contour and its intensity will decrease after the dilated convolution.
The vacancy in the kernel of the dilated convolution causes the loss of information continuity. Thus, two plain convolution blocks are added to increase the transition between pixels.
In addition, the number of channels is reduced for parameter reduction and bilinear upsampling is once again used for size adjustment. 
Consequently, the contour-sensitive maps, $i.e.,\{H\}$, which contain context information, are gained.

\subsection{Complementary Feature Fusion}
\label{sec:CFF}

In view of the disparity between multi-level features, we utilize \textbf{CFF} to make up the details of high-level features and suppress the noise of low-level features. To be specific, it has complementary function to restore details of contours and remove clutter of non-occluded pixels.

In \textbf{CFF}, it is necessary to fuse three disparate streams of features, $i.e.,$ spatial-refined maps $\{S_1,S_2,S_3\}$, contour-sensitive maps $\{H\}$ and backbone decoder maps $\{D\}$. Among them, $\{S_1,S_2,S_3\}$ reflect the edge locations and structure information of low-level features, $\{H\}$ represent the contour information of high-level features, $\{D\}$ express the hierarchical and global information of the entire decoder. 
There exists precise location information in the fused low semantic features, which is able to restore details of $\{H\}$. On this account, the locations of the occluded contours are more accurate through \textbf{CFF}.
Simultaneously, the fused high semantics features imply the relationship between pixels and the contours around them. Context information is able to determine whether the pixel is noise or a contour pixel, thus noise of $\{S_1,S_2,S_3\}$ is suppressed. 
We utilize convolution blocks to extract features and reduce channels. Then a batch normalization layer is in service to balance features at different levels. In the end, a $ReLU$ \cite{glorot2011deep} is taken on to amend the elements that do not conform to the range of the edge. 
Consequently, fused features from different levels promote network to select element-wise useful information. $\{F\}=\textbf{CFF}(\{S_1,S_2,S_3\},\{H\},\{D\})$ 
are employed to denote the complementary fusion maps.

\subsection{Attention Weight Response}
\label{sec:AWR}

Occlusion edge is of high semantic, hence weight-averaged elements are unreasonable in feature maps. Specifically, it is necessary to train a module with semantic information, which locates the contours and uses the weight response of attention mechanism to suppress the noise of non-edge pixels.

Considering that the contour-sensitive maps $\{H\}$ have large receptive fields and are sensitive to the entire contour, we utilize them as weight response of attention mechanism to refine the complementary fusion maps $\{F\}$.
The attention weight maps $\{C\}=\textbf{AWR}(\{H\})$ are proposed to further refine $\{F\}$, which roughly locate the contour position to suppress the texture around it. 
By employing convolution blocks to acquire contour context from \textbf{CSA}, we first obtain a high response feature map with single channel. A $Sigmoid$ is then used to convert the elements in feature maps into corresponding weight values. To assign higher weight for the contour regions and lower weight for the non-occluded edges, an element-wise product operation with $\{F\}$ and $\{C\}$ is conducted shown in Fig.\ref{Fig:modules}(b). As a result, the resulting feature maps $\{R\}=\{F\}\otimes\{W\}$ not only increase the contour strength, but also weaken noise around the occlusion edges.


\section{EXPERIMENTS}

In this section, we present a comprehensive evaluation of the proposed method.
Experimental results are reported on two challenging datasets, i.e. PIOD dataset \cite{Wang2016DOC} and BSDS Ownership dataset \cite{Ren2006Figure}. 
Several ablation studies are further presented on the PIOD dataset.



\subsection{Experimental Settings}


PIOD dataset has 9,175 training images and 925 testing images. Each image in the dataset has an object instance edge map and the corresponding occlusion orientation map. BSDS ownership dataset has 100 training images and 100 testing images. 
Following \cite{Wang2018DOOBNet}, all images are randomly cropped to 320 $\times$ 320 during training. And we operate on an input image at its original size during testing.

{\bf{Implementation Details:}}
Our CCENet is implemented using Caffe \cite{Jia2014Caffe} and runs on a single NVIDIA GeForce GTX 1080 Ti. The whole network is fine-tuned from an initial pretrained $Res50$ model. All convolution layers added are initialized with the "msra" \cite{He2015Delving}.

\begin{table}[t]
\small
\caption{EPR results on PIOD (left) and BSDS ownership dataset (right). \ding{172}-\ding{176} represent SRF-OCC \cite{Teo2015Fast}, DOC-HED \cite{Wang2016DOC}, DOC-DMLFOV \cite{Wang2016DOC} and DOOBNet \cite{Wang2018DOOBNet} and our CCENet, respectively. Red bold type indicates the best performance, blue bold type indicates the second best performance. (the same below)
}
\renewcommand\arraystretch{1.3}
\begin{center}
\begin{tabular}{ l | l  l  l | l  l  l}
\hline
\emph{ }  &ODS &OIS &AP &ODS &OIS &AP\\
\hline
\ding{172}   &$.345$  &$.369$ &$.207$ &$.511$ &$.544$ &$.442$\\
\hline
\ding{173}   &$.509$  &$.532$ &$.468$ &$\textbf{{\color{red}.658}}$ &$\textbf{{\color{red}.685}}$ &$\textbf{{\color{red}.602}}$\\
\ding{174}   &$.669$  &$.684$ &$.677$ &$.579$ &$.609$ &$.519$\\
\hline
\ding{175}   &$\textbf{{\color{blue}.736}}$  &$\textbf{{\color{blue}.746}}$ &$\textbf{{\color{blue}.723}}$ &$.647$ &$.668$ &$.539$\\
\ding{176}   &$\textbf{{\color{red}.744}}$  &$\textbf{{\color{red}.758}}$ &$\textbf{{\color{red}.767}}$ &$\textbf{{\color{blue}.649}}$ &$\textbf{{\color{blue}.673}}$ &$\textbf{{\color{blue}.549}}$\\
\hline

\end{tabular}
\label{tab:EPR}
\end{center}
\end{table}

\begin{table}[t]
\small
\caption{OPR results on PIOD (left) and BSDS ownership dataset (right). $\dagger$ refers to GPU running time.
}
\renewcommand\arraystretch{1.3}
\begin{center}
\begin{tabular}{ p{0.3cm} | p{0.4cm} p{0.4cm} p{0.4cm} p{0.9cm} | p{0.4cm} p{0.4cm} p{0.4cm} p{0.9cm}}
\hline
\emph{ }  &ODS &OIS &AP &FPS &ODS &OIS &AP &FPS\\
\hline
\ding{172}   &$.268$ &$.286$ &$.152$ &$0.018$ &$.419$ &$.448$ &$.337$ &$0.018$\\
\hline
\ding{173}   &$.460$ &$.479$ &$.405$ &$18.3\dagger$ &$.522$ &$.545$ &$.428$ &$19.6\dagger$\\
\ding{174}   &$.601$ &$.611$ &$.585$ &$18.9\dagger$ &$.463$ &$.491$ &$.369$ &$21.1\dagger$\\
\hline
\ding{175}   &$\textbf{{\color{blue}.702}}$ &$\textbf{{\color{blue}.712}}$ &$\textbf{{\color{blue}.683}}$ &$26.7\dagger$ &$\textbf{{\color{blue}.555}}$ &$\textbf{{\color{blue}.570}}$ &$\textbf{{\color{blue}.440}}$ &$25.8\dagger$\\
\ding{176}   &$\textbf{{\color{red}.712}}$ &$\textbf{{\color{red}.722}}$ &$\textbf{{\color{red}.724}}$ &$28.3\dagger$ &$\textbf{{\color{red}.582}}$ &$\textbf{{\color{red}.599}}$ &$\textbf{{\color{red}.452}}$ &$27.2\dagger$\\
\hline

\end{tabular}
\label{tab:OPR}
\end{center}
\end{table}

{\bf{Hyper-parameters:}}
The hyper-parameters include: mini-batch size (5), iter size (3), learning rate (3e-5), momentum (0.9), weight decay (0.0002),
$\lambda$ (0.5) in Eq.\eqref{Eq:loss_all}.
The number of training iterations for PIOD dataset is 30,000 (divide learning rate by 10 after 20,000 iterations) and BSDS ownership dataset is 5,000 (divide learning rate by 10 after every 2,000 iterations). 

\subsection{Occlusion Edge Detection Performance}


\quad\;\;{\bf{Evaluation Criteria.}}
To evaluate object edge detection and occlusion orientation detection, we calculate precision and recall of the predicted edges and orientations (described as EPR and OPR) by performing three standard metrics: fixed contour threshold (ODS), best threshold of each image (OIS) and average precision (AP).
For both cases, we use these three metrics to evaluate the performance.
Notably, with regard to OPR, we compute recall only at the correctly detected edge pixels.



\begin{figure}[t]
\centerline{\psfig{figure=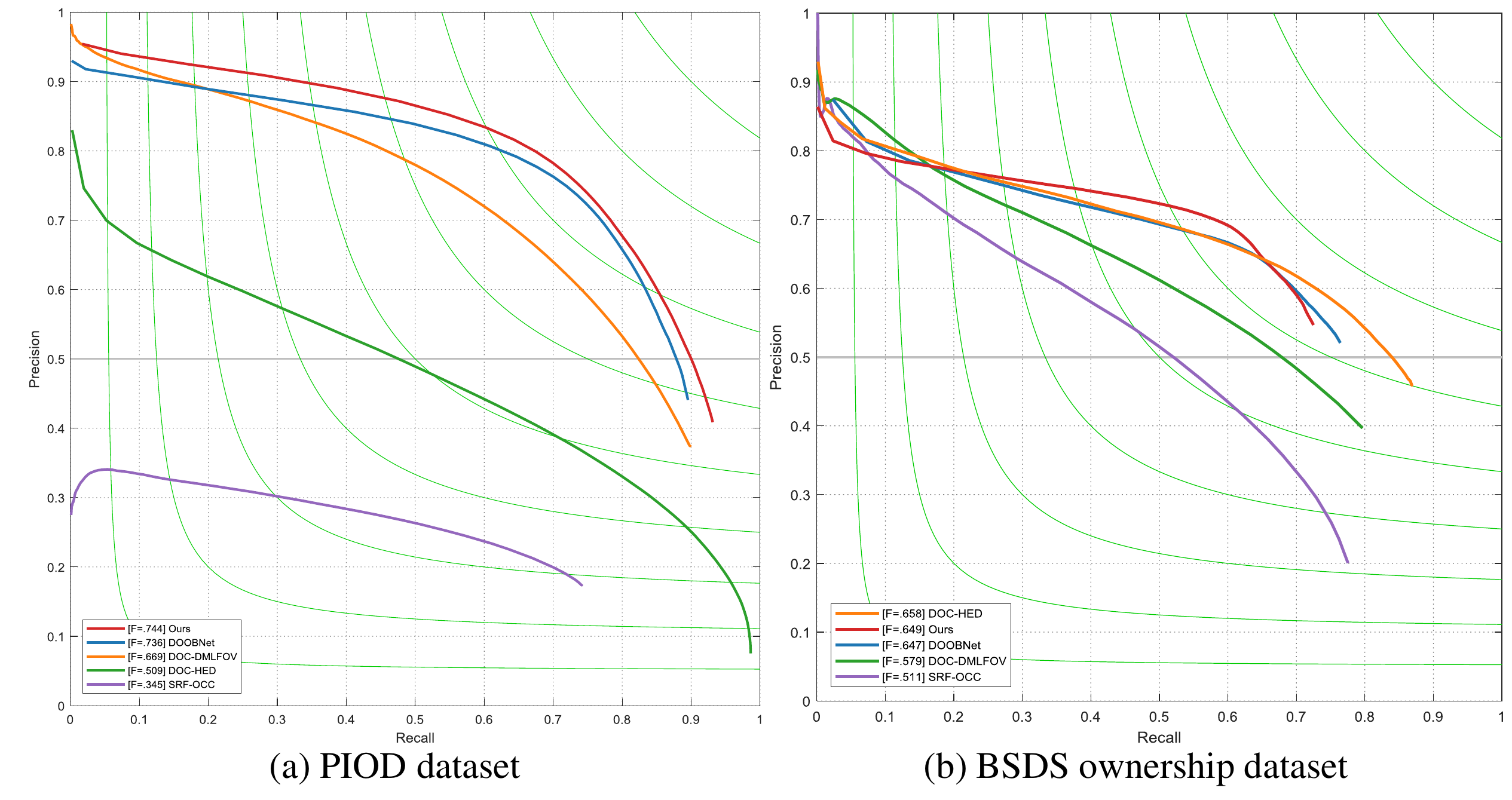,width=3.4in}}
\caption{
EPR results on two datasets.
}
\label{Fig:EPR}
\end{figure}

\begin{figure}[t]
\centerline{\psfig{figure=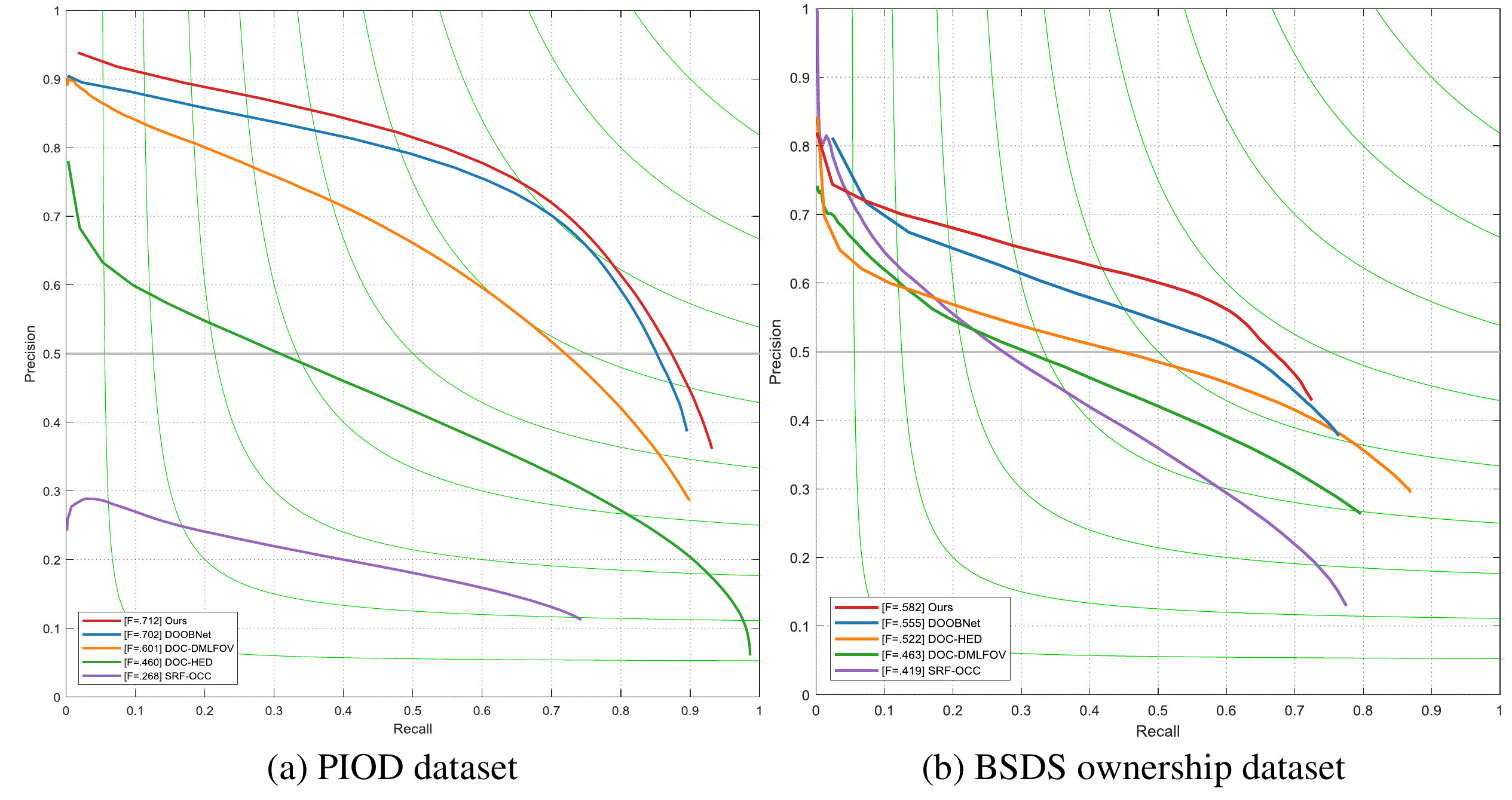,width=3.4in}}
\caption{
OPR results on two datasets.
}
\label{Fig:OPR}
\end{figure}

{\bf{Quantitative Analysis.}}
We evaluate our approach against numerous state-of-the-art approaches, i.e., SRF-OCC \cite{Teo2015Fast}, DOC-HED \cite{Wang2016DOC}, DOC-DMLFOV \cite{Wang2016DOC} and DOOBNet \cite{Wang2018DOOBNet}. 
EPR results on the PIOD dataset are shown in Table \ref{tab:EPR}.
The proposed approach performs favorably against other state-of-art methods in all metrics. 
Our approach utilizes \textbf{CSA} and \textbf{AWR} to further enhance the response of the contours, resulting in detecting less noise.
As shown in Fig.\ref{Fig:EPR}(a), our approach achieves prominent high precision when recall is at a low value since CCENet is sensitive to contour regions. 
OPR results are exhibited in Table \ref{tab:OPR}. Our approach performs the best, achieving ODS=.712, because CCENet rationally utilize \textbf{CFF} to restore details and suppress noise of fused features. It is 1.0\%, 11.1\% and 25.2\% higher than DOOBNet, DOC-DMLFOV and DOC-HED, respectively. As shown in Fig.\ref{Fig:OPR}(a), our method possesses low false inspection, on account of augmenting the role of the high semantic layers.

We also train and test our approach on the BSDS ownership dataset.
The annotations of the dataset are not all occlusion edges, leading to missed inspection of edges within the object of our approach, which is reasonable in occlusion edge detection yet. 
On account of EPR, as illustrated in Table \ref{tab:EPR} and Fig.\ref{Fig:EPR}(b), CCENet is slightly lower than DOC-HED, yet having high performance when recall $<$ 0.7. Specific to OPR, presented in Table \ref{tab:OPR} and Fig.\ref{Fig:OPR}(b), CCENet outperforms DOOBNet significantly by a margin of 2.7\% ODS.

\begin{figure}[t]
\centerline{\psfig{figure=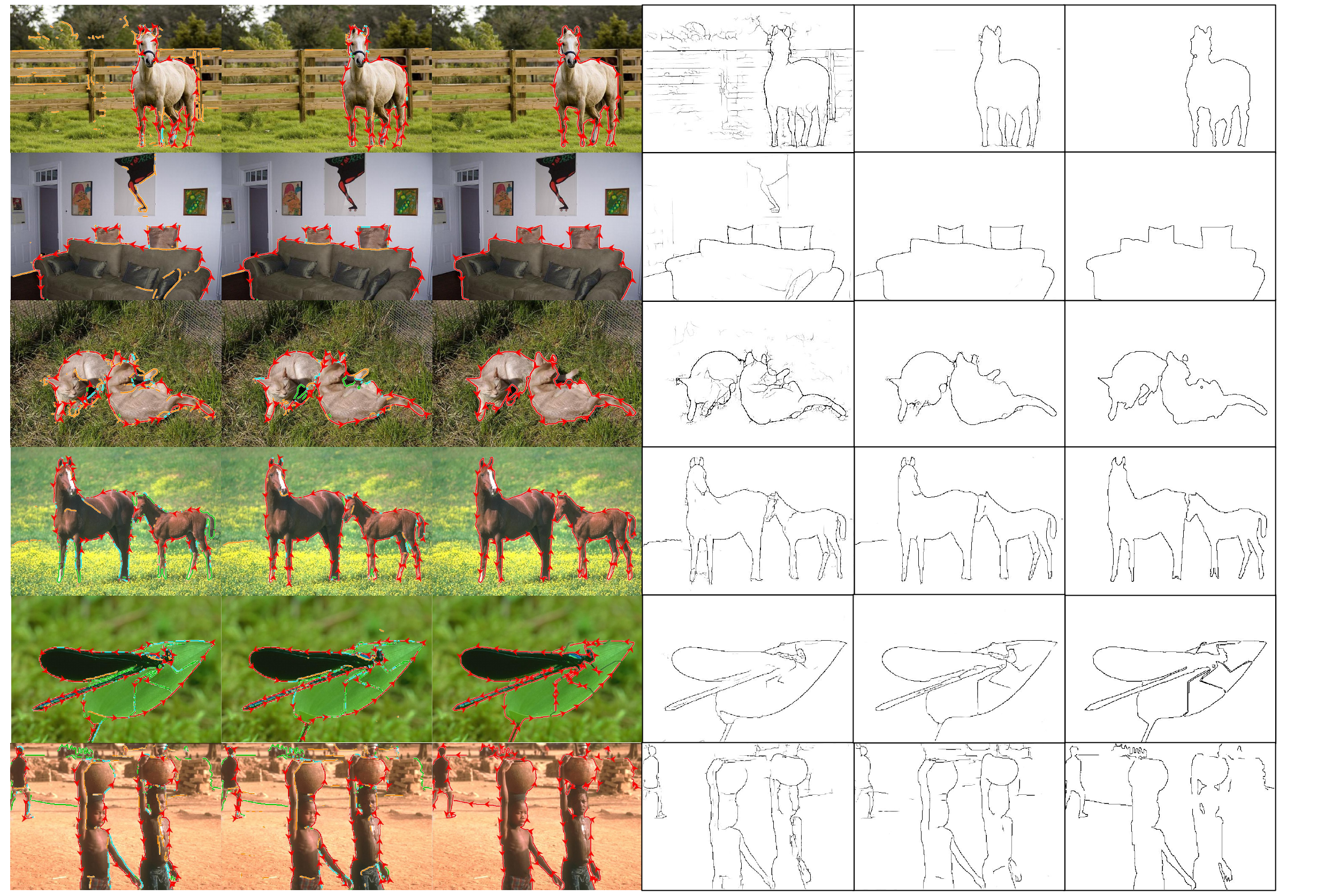,width=3.4in}}
\caption{
Example results on PIOD (first three rows) and BSDS ownership dataset (last three rows). Visualization results of DOOBNet, CCENet and ground truth (first three columns): "red" pixels with arrows: correctly labeled occlusion boundaries; "cyan": correctly labeled boundaries but mislabeled occlusion; "green": false negative boundaries; "orange": false positive boundaries. Corresponding object boundaries (last three columns). (Best viewed in color)
}
\label{Fig:quality}
\end{figure}

{\bf{Qualitative Analysis.}}
A set of qualitative results on two datasets
are exhibited in Fig.\ref{Fig:quality}. 
The leaves of the trees and the pattern on the picture represent low-level texture. They are not identified as contours thanks to the role of the high semantic layers. The extremely slender limbs of the insect are accurately extracted due to the details provided by the low semantic layers.

\subsection{Ablation Study}

We conduct various ablation studies to analyze the details of our approach on the PIOD dataset.

{\bf{Ablation for Spatial Feature Refinement.}}
As 
illustrated
in Sec.\ref{sec:SFR}, 
DOOBNet does not take advantage of side outputs adequately of lower-level layers, leading to the challenge of inaccurate locations of details. Hence, \textbf{SFR} is designed to make full use of the detailed information. The gap of the comparison 
in Table \ref{tab:Ablation} explains the effect of the \textbf{SFR}.

{\bf{Ablation for Contour Sensitivity Augmentation.}}
We expect \textbf{CSA} to supply adequate receptive field.
It acquires affluent context information between pixels on the contour, which is sensitive to the whole contour. Effect of \textbf{CSA} is presented in Table \ref{tab:Ablation}.

{\bf{Ablation for Complementary Feature Fusion.}}
Considering features of various levels, \textbf{CFF} is proposed to fuse the features of low-level and high-level outputs. In contrast with 
simple concatenation operations, this yields 0.2\% ODS, 0.3\% OIS, 1.3\% AP of EPR, as shown in Table \ref{tab:Ablation}.

{\bf{Ablation for Attention Weight Response.}}
We design \textbf{AWR} in order to promote the detection performance further. By utilizing weight maps, the fusion feature maps
are
re-weighted. Consequently, the attention weight maps strengthen the contours and weaken noise around the semantic edges simultaneously. 
This gives the gains of 0.2\% ODS, 0.4\% OIS, 1.2\% AP of EPR, as shown in Table \ref{tab:Ablation}.

\section{CONCLUSION}

In this paper, 
a novel CCENet is proposed
for occlusion edge detection. We exploit low-level and high-level information from side outputs to extract edge spatial and contour context features, aiming to obtain accurate locations and perceive the entire contour. Complementary Feature Fusion (\textbf{CFF}) is designed to aggregate features from various levels and combine the advantages of each other. Attention Weight Response (\textbf{AWR}) utilizes the weight response of attention mechanism to strengthen the contours and suppress the noise. 
Extensive experimental results show that the proposed algorithm performs
favorably against the state-of-the-art methods with significant margins.

\begin{table}[t]
\footnotesize
\caption{Ablation results on the PIOD dataset (left: EPR results, right: OPR results). Our baseline DOOBNet is in the first row, taken as the initial structure. The symbol "$+$" means adding the module to the previous network structure.
}
\renewcommand\arraystretch{1.3}
\begin{center}
\begin{tabular}{ l | p{0.4cm} p{0.4cm} p{0.4cm} | p{0.4cm} p{0.4cm} p{0.4cm}}
\hline
\emph{Methods}  &ODS &OIS &AP &ODS &OIS &AP\\
\hline
DOOBNet            &$.736$ &$.746$ &$.723$  &$.702$ &$.712$ &$.683$\\
\hline
$+$\textbf{SFR}      &$.738$ &$.748$ &$.731$  &$.704$ &$.713$ &$.691$\\
$+$\textbf{CSA}              &$.740$ &$.751$ &$.742$  &$.707$ &$.716$ &$.703$\\
\hline
$+$\textbf{SFR}$+$\textbf{CSA}(CONCAT)   &$.741$ &$.751$ &$.746$  &$.707$ &$.717$ &$.708$\\
$+$\textbf{SFR}$+$\textbf{CSA}(\textbf{CFF})      &$\textbf{{\color{blue}.742}}$ &$\textbf{{\color{blue}.754}}$ &$\textbf{{\color{blue}.755}}$  &$\textbf{{\color{blue}.710}}$ &$\textbf{{\color{blue}.719}}$ &$\textbf{{\color{blue}.719}}$\\
\hline
$+$\textbf{SFR}$+$\textbf{CSA}(\textbf{CFF})$+$\textbf{AWR}  &$\textbf{{\color{red}.744}}$ &$\textbf{{\color{red}.758}}$ &$\textbf{{\color{red}.767}}$  &$\textbf{{\color{red}.712}}$ &$\textbf{{\color{red}.722}}$ &$\textbf{{\color{red}.724}}$\\
\hline
\end{tabular}
\label{tab:Ablation}
\end{center}
\end{table}



\bibliographystyle{IEEEbib}
\bibliography{icme2019template}

\end{document}